\documentclass{article} 

\usepackage{iclr2026_conference,times}

\makeatletter
\renewenvironment{abstract}{%
    \vskip.075in\centerline{\large\sc Abstract}\vspace{0.5ex}%
}{%
    \par\vskip 1ex%
}
\makeatother

\usepackage{amsmath,amsfonts,bm}

\def\eqref#1{equation~\ref{#1}}

\def\1{\bm{1}}

\DeclareMathAlphabet{\mathsfit}{\encodingdefault}{\sfdefault}{m}{sl}
\SetMathAlphabet{\mathsfit}{bold}{\encodingdefault}{\sfdefault}{bx}{n}

\usepackage{hyperref}
\usepackage{url}
\usepackage{graphicx}
\usepackage[most]{tcolorbox}
\usepackage{wrapfig}
\usepackage{booktabs}
\usepackage{xcolor}

\usepackage{arabtex}
\usepackage{utf8}

\title{When Alignment Hurts: Decoupling Representational Spaces in Multilingual Models}

\author{Ahmed Elshabrawy\textsuperscript{1}\thanks{work done during internship at NICT, Japan} \, Hour Kaing\textsuperscript{2}\, Haiyue Song\textsuperscript{2}\, Alham Fikri Aji\textsuperscript{1} \, Hideki Tanaka\textsuperscript{2}\\ \textbf{Masao Utiyama\textsuperscript{2} \, Raj Dabre\textsuperscript{3}}\thanks{work done during tenure at NICT, Japan and unrelated to current position at Google.} \\
\textsuperscript{1}MBZUAI \,\, \textsuperscript{2}NICT, Japan \,\, \textsuperscript{3} IIT Madras \\
\texttt{ahmed.elshabrawy@mbzuai.ac.ae} \\
\texttt{raj.dabre@cse.iitm.ac.in}
}

\iclrfinalcopy 
\begin{document}

\maketitle


\begin{abstract}
\begin{quote}
    Alignment with high-resource standard languages is often assumed to aid the modeling of related low-resource varieties. We challenge this assumption by demonstrating that excessive representational entanglement with a dominant variety, such as Modern Standard Arabic (MSA) in relation to Arabic dialects, can actively hinder generative modeling. We present the first comprehensive causal study of this phenomenon by analyzing and directly intervening in the internal representation geometry of large language models (LLMs). Our key contribution is an online variational probing framework that continuously estimates the subspace of the standard variety during fine-tuning, enabling projection-based decoupling from this space. While our study uses Arabic as a case due to its unusually rich parallel resources across 25 dialects, the broader motivation is methodological: dialectal MT serves as a controlled proxy for generative tasks where comparable multi-variety corpora are unavailable. Across 25 dialects, our intervention improves generation quality by up to +4.9 chrF++ and +2.0 on average compared to standard fine-tuning, despite a measured tradeoff in standard-language performance. These results provide causal evidence that subspace dominance by high-resource varieties can restrict generative capacity for related varieties. More generally, we unify geometric and information-theoretic probing with subspace-level causal interventions, offering practical tools for improving generative modeling in closely related language families and, more broadly, for controlling representational allocation in multilingual and multi-domain LLMs. Code will be released.
\end{quote}
\end{abstract}

\section{Introduction}

Large Language Models (LLMs) have achieved remarkable progress in multilingual Natural Language Understanding (NLU) and Generation (NLG) tasks \citep{brown2020language, chowdhery2022palm, scao2022bloom, aryabumi2024aya23openweight}. Beyond English, these models show strong cross-lingual transfer, enabling low-resource varieties to benefit from related high-resource languages \citep{hu2020xtreme, conneau2020unsupervised, xue-etal-2021-mt5}. 

A less understood question, however, is whether closer alignment with a dominant, high-resource variety always benefits related low-resource ones. Dialects provide a natural test case: they are linguistically distinct, socially important, yet often heavily entangled with their standardized counterpart in both data and models. Arabic exemplifies this dynamic, where Modern Standard Arabic (MSA) dominates pretraining resources while dozens of dialects remain underrepresented and underperform on benchmarks \citep{kantharuban-etal-2023-quantifying}. Similar dynamics arise in other orthographically and lexically close pairs such as Czech–Slovak or Swedish–Icelandic. Understanding representational interactions in such settings is crucial for inclusive generative modeling.

This paper challenges the assumption that alignment with a high-resource standard is always beneficial. Using Arabic dialects as a case study, chosen for their unusually rich parallel resources across 25 varieties \citep{bouamor-etal-2018-madar}, we show that excessive representational entanglement with MSA hinders generative performance. Since parallel corpora for other generative tasks across dialects are scarce, we focus on machine translation as a controlled proxy for dialect-sensitive generation.

Our study proceeds in two stages. First, we analyze how LLMs internally represent MSA and dialects, revealing that stronger generative performance correlates with greater representational separability from MSA. Second, we move from analysis to intervention: we introduce an \textbf{online variational probing} framework that continuously estimates the subspace of the high-resource standard during fine-tuning, enabling a novel subspace decoupling strategy. This causal intervention promotes orthogonal representations and improves generative capacity for dialects.

\begin{table}[ht]
\centering
\caption{Sample of 5-way parallel sentences meaning "\colorbox{yellow!30}{How much} does \colorbox{blue!30}{the breakfast} \colorbox{green!30}{cost}?" in 5 different varieties of Arabic from the MADAR 26 corpus \citep{bouamor-etal-2018-madar}. The yellow highlights the interrogative element (roughly “how much”), the green (when present) highlights the explicit cost word, and the blue highlights the breakfast term.}

\setcode{utf8}
\begin{tabular}{|c|c|c|}
\hline
\textbf{Dialect} & \textbf{Arabic} & \textbf{Transliteration (Buckwalter)} \\
\hline
Modern Standard Arabic & 
\colorbox{blue!30}{\RL{الإفطار؟}}~\colorbox{green!30}{\RL{تكلفة}}~\colorbox{yellow!30}{\RL{كم}} 
& \colorbox{yellow!30}{kam}~\colorbox{green!30}{taklifaT}~\colorbox{blue!30}{al-'ifTar?} \\
\hline
Egyptian Arabic & 
\colorbox{blue!30}{\RL{الفطار؟}}~\colorbox{yellow!30}{\RL{بكام}} 
& \colorbox{yellow!30}{bkam}~\colorbox{blue!30}{al-fiTar?} \\
\hline
Levantine Arabic & 
\colorbox{blue!30}{\RL{الترويقة؟}}~\colorbox{green!30}{\RL{حق}}~\colorbox{yellow!30}{\RL{أدي}} 
& \colorbox{yellow!30}{'addi}~\colorbox{green!30}{Haq}~\colorbox{blue!30}{al-tarwiqa?} \\
\hline
Gulf Arabic & 
\colorbox{blue!30}{\RL{الريوق؟}}~\colorbox{yellow!30}{\RL{بكم}} 
& \colorbox{yellow!30}{bkam}~\colorbox{blue!30}{al-riyooq?} \\
\hline
Maghrebi Arabic & 
\colorbox{blue!30}{\RL{فطور الصباح؟}}~\colorbox{yellow!30}{\RL{بقداش}} 
& \colorbox{yellow!30}{bqaddash}~\colorbox{blue!30}{fuToor al-SabaaH?} \\
\hline
\end{tabular}

\label{tab:dia-eg}
\end{table}

Applied to 25 Arabic dialects, our approach yields consistent improvements over standard fine-tuning, up to +4.9 chrF++ on individual dialects and +2.0 on average, while trading off some performance in MSA generation. More broadly, our findings provide the first causal evidence that representational dominance by high-resource standards can limit generative modeling in closely related varieties. 

\textbf{Contributions.} 
\begin{itemize}
    \item We present the first large-scale representational analysis of dialects in generative LLMs, unifying geometric and information-theoretic probing. 
    \item We introduce a novel online probing-based subspace decoupling method that improves generative performance for underrepresented varieties. 
    \item We empirically demonstrate consistent gains across 25 dialects, highlighting implications for related language families where orthographic and lexical similarity creates similar entanglement.
\end{itemize}

\section{Related Works}
This work investigates how LLMs internally allocate representational capacity across closely related language varieties, with a focus on Arabic dialects as a natural case study.

\paragraph{Multilingualism in Large Language Models.} 
Recent studies have analyzed how multilingual LLMs encode language-specific knowledge. For example, \citet{wang2024sharing} and \citet{kojima2024multilingual} explore neuron sharing and language-specific activations, showing that subtle modifications can alter generation in particular languages. Our perspective differs: rather than focusing on neuron-level behavior, we ask whether dialects remain representationally distinct from their standardized counterpart and how this distinction (or entanglement) affects generative performance. This question is not limited to Arabic, but applies broadly to orthographically and lexically similar pairs with a resource imabalance.

At the representational level, \citet{chang-etal-2022-geometry} show that languages occupy distinct subspaces in encoder-only models, while \citet{shah-etal-2024-correlations} link geometric differences to cross-lingual transfer. We extend these insights to large generative models, showing that the degree of subspace separability between varieties correlates with downstream generation quality. Relatedly, \citet{nigatu-etal-2023-less} find that models struggle to capture dialectal nuances; our results both confirm this for recent LLMs and provide causal evidence that mitigating representational entanglement improves performance.

\paragraph{Information-Theoretic Probing.}
Information-theoretic probes have been used to study how linguistic signals emerge during pretraining \citep{voita-titov-2020-information, muller-eberstein-etal-2023-subspace}. Building on this, we introduce probes not just for analysis but as part of training: our ``dialect probes’’ continuously estimate the dominant standard-language subspace during fine-tuning, enabling us to directly intervene by penalizing entanglement. This extends probing from a diagnostic tool to a mechanism for causal representational control.

\paragraph{Dialectal and Low-Resource NLP.}
Dialectal variation presents a persistent challenge for generative modeling. Prior work has documented large performance gaps as dialects deviate from their standardized counterpart \citep{kantharuban-etal-2023-quantifying, ziems-etal-2023-multi}. For Arabic, evaluation resources such as AraBench \citep{sajjad-etal-2020-arabench} and MADAR \citep{bouamor-etal-2018-madar} have been developed, and recent studies examine MT and NLG across varieties \citep{kadaoui-etal-2023-tarjamat, nagoudi-etal-2023-dolphin}. Our work departs from these by focusing not on resource creation or evaluation but on how dialects are internally represented in LLMs and how interventions on representational subspaces can improve generative capacity. While Arabic provides a uniquely rich testbed, the implications extend to other under-resourced language varieties that share high orthographic and lexical overlap with a dominant variety.

\section{Background: Arabic Dialects}\label{sec:background}

A significant challenge in developing truly multilingual models lies in handling closely-related language varieties, which often exist in a state of resource imbalance with a dominant, high-resource standard language. This scenario is not merely a linguistic curiosity but poses fundamental problems for model representation learning. We investigate this challenge through the lens of Arabic, which provides an ideal testbed due to its distinct diglossia. The language ecosystem consists of Modern Standard Arabic (MSA), a high-resource variety used in formal contexts, and numerous low-resource Dialectal Arabic (DA) varieties that are the primary spoken languages but lack substantial textual corpora. The availability of the MADAR corpus \citep{bouamor-etal-2018-madar}, a unique resource with parallel sentences across 25 dialects, enables a controlled study of this phenomenon, which is not feasible for many other language families with similar dialectal diversity.

\begin{wrapfigure}{r}{0.4\textwidth}
    \vspace{-10pt} 
    \includegraphics[width=\linewidth]{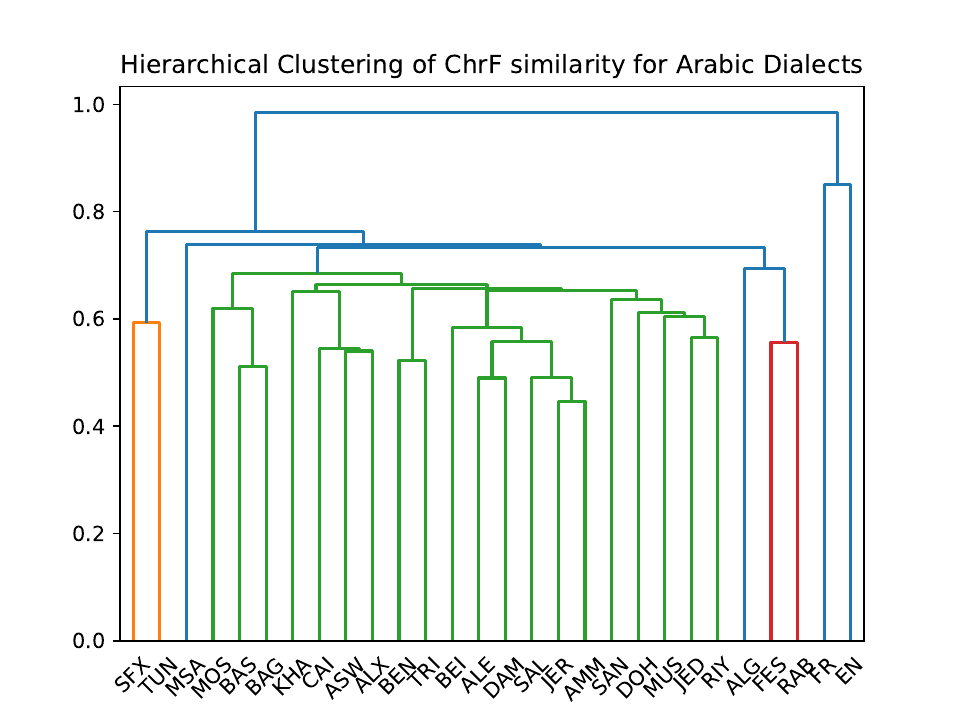} 
    \caption{Hierarchical clustering of Arabic varieties based on character-level distance (chrF++). Dialects cluster geographically and exhibit significant separation from the high-resource Modern Standard Arabic (MSA), quantifying their surface-level dissimilarity.}
    \label{fig:dia-cluster}
    \vspace{-10pt} 
\end{wrapfigure}

The divergence between MSA and DA is substantial, spanning lexical, syntactic, and morphological levels, as illustrated by the parallel translations in Table~1. To quantify this, we perform hierarchical clustering based on character-level similarity (chrF++ score, \citep{popovic-2015-chrf}) in Figure~\ref{fig:dia-cluster}. The analysis reveals that dialects form distinct, geographically-correlated clusters that are representationally distant from MSA. This dissimilarity presents a concrete technical obstacle, starting at the tokenization layer. As detailed in Appendix~\ref{sec:tokenizers}, standard model tokenizers trained predominantly on high-resource data yield suboptimal segmentations for dialectal words. This inefficiency leads to higher computational costs and hinders the model's ability to learn long-range dependencies, unfairly disadvantaging low-resource varieties before any deeper processing occurs \citep{ali-etal-2024-tokenizer}.

Our goal is to understand how the representational dominance of a high-resource language like MSA affects a model's ability to generate text in closely-related, low-resource dialects. Due to the scarcity of parallel data for diverse generative tasks, we utilize machine translation (MT) from MSA to the 25 DA varieties as a controlled testbed for generation. This task allows us to probe the model's capacity to produce dialect-specific outputs while controlling for semantic content. While our empirical study is grounded in Arabic, the core challenge is generalizable. The insights derived are relevant to other language families with similar orthographic overlap and resource asymmetry, such as Czech and Slovak, or the spectrum of Scandinavian languages, where models must learn to navigate the subtle yet critical distinctions between high- and low-resource variants.
\section{Methodology}
We present a methodology designed to first diagnose and then causally intervene in the representational geometry of multilingual models. Our approach uses a controlled generative task to probe model capabilities, analyzes the underlying representations through geometric and information-theoretic lenses, and introduces a novel training technique to mitigate representational entanglement.

\subsection{Task Formulation: Dialectal Rewriting as a Generative Testbed}
To create a controlled environment for studying dialectal generation, we formulate a task of \textbf{Dialectal Machine Translation} (DiaMT). Given a sentence in the high-resource standard variety (MSA), the model's objective is to generate the semantically equivalent sentence in a target low-resource dialect. This task serves as a valuable proxy for more general conditional generation, allowing us to precisely measure a model's ability to manipulate linguistic style while preserving meaning. This controlled setup is necessitated by the lack of comprehensive parallel corpora for other generative tasks (e.g., summarization, open-ended dialogue) across the 25 dialects. We employ zero-shot inference using a simple instructional prompt, as shown below.

\begin{wrapfigure}[25]{r}{0.5\textwidth}
\centering
\begin{minipage}{0.48\textwidth}
    \begin{tcolorbox}[enhanced, sharp corners, colback=white, colframe=blue!40,
    title=\textbf{Prompt Format}, fonttitle=\bfseries, boxrule=1pt, left=6pt, right=6pt, top=6pt, bottom=6pt]
    Rewrite the following from Modern Standard Arabic to the dialect of the Cairo city dialect.
    \vspace{1em}
    \textbf{MSA phrase:} \{\{MSA Sentence\}\}
    
    \textbf{Cairo phrase:}
    \end{tcolorbox}
\end{minipage}
\hfill
\begin{minipage}{0.48\textwidth}
    \includegraphics[width=\linewidth]{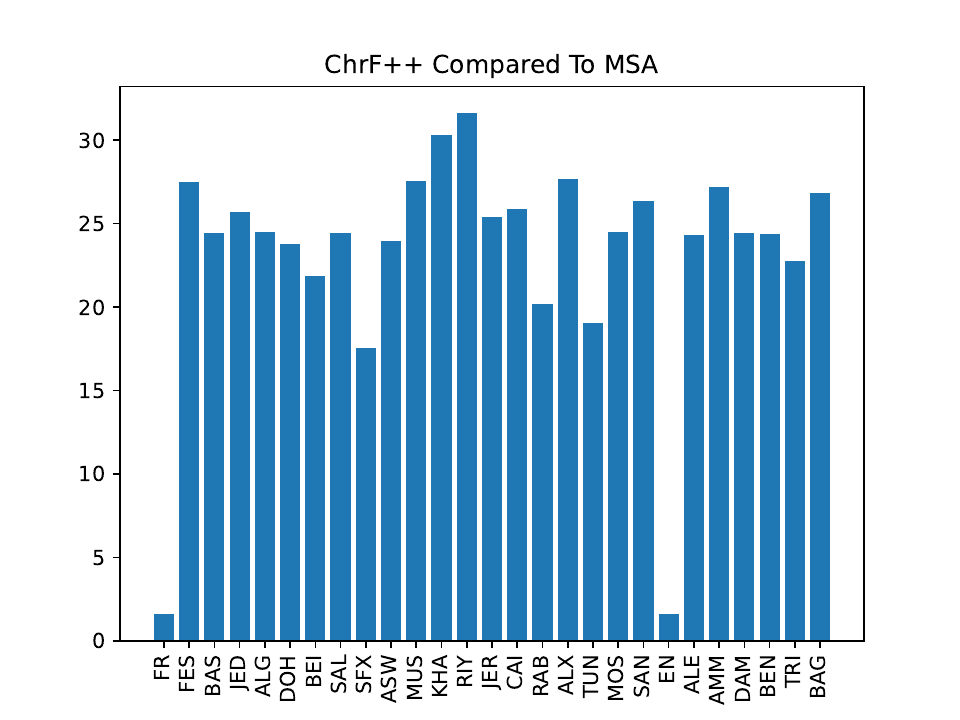}
    \caption{Average character-level similarity (chrF++) between dialectal sentences and their MSA counterparts, highlighting the high surface overlap that makes this a challenging generation task.}
    \label{fig:chr-msa}
\end{minipage}
\end{wrapfigure}

For our causal experiments (Sec.~\ref{sec:decouple}), we fine-tune models on a bidirectional rewriting objective (MSA $\leftrightarrow$ dialect). This prevents the model from simply degrading its high-resource MSA representations to favor dialects, a potential confounding factor in unidirectional training. By preserving MSA capabilities, we ensure a fairer assessment of subspace dynamics and the true impact of our intervention on dialectal generation.

\subsection{Quantifying Performance and Representational Geometry}
\paragraph{Evaluation.} We quantify generation quality using chrF++ \citep{popovic-2015-chrf}, a character n-gram F-score. Its character-level nature is well-suited for the morphological richness of Arabic and is robust to minor lexical variations common between dialects. We acknowledge that, like many automated metrics, chrF++ cannot fully capture the nuances of "dialectness," especially given the high lexical overlap between Arabic varieties (Fig.~\ref{fig:chr-msa}). However, in the absence of better-suited metrics for this specific cross-dialectal evaluation, it serves as a reliable indicator of generative accuracy.

\paragraph{Representational Geometry.} To understand \textit{how} models represent dialects, we analyze their internal geometry. We measure the \textbf{Geometric Separability} between sentence representations using L2 and cosine distance, anchoring all comparisons to MSA representations. This allows us to quantify how distinct dialectal representations are from the high-resource standard. Furthermore, we compute \textbf{Subspace Angles (SSA)} \citep{muller-eberstein-etal-2023-subspace} to measure the alignment between subspaces corresponding to different dialects. Smaller angles indicate greater alignment. This allows us to track how fine-tuning and our proposed interventions reshape the model's internal organization of linguistic information.

\subsection{Information-Theoretic Probing}
\label{sec:probing}
To complement the geometric analysis, we employ an information-theoretic variational linear probe \citep{voita-titov-2020-information, muller-eberstein-etal-2023-subspace}. The probe is a sparsity-regularized classifier trained to identify a dialect from token-level representations. The resulting negative cross-entropy provides a tight lower bound on the mutual information $I(\mathbf{h}^{(\ell)}; Y)$ between a model's hidden states and the dialect identity. This allows us to quantify how easily dialect-specific information can be linearly decoded from the model's representations, layer by layer, and how this changes during training. Further details are in Appendix~\ref{sec:probing-detailed}.

\subsection{Causal Intervention: Online Subspace Decoupling}
\label{sec:decouple}
To test the hypothesis that representational entanglement with a high-resource language harms low-resource generation, we introduce a novel training method: \textbf{Online Subspace Decoupling}. This method acts as a causal intervention by actively discouraging dialectal representations from overlapping with the MSA subspace during fine-tuning.

The procedure is as follows:
\begin{enumerate}
    \item \textbf{Identify MSA Subspace:} We train a variational linear probe (as in Sec.~\ref{sec:probing}) to distinguish MSA from all other dialects. We then use Singular Value Decomposition (SVD) on the learned probe weights to extract an orthonormal basis $\mathbf{U}_{\text{MSA}}$ for the MSA subspace and form its projection matrix: $\mathbf{P}_{\text{MSA}} = \mathbf{U}_{\text{MSA}} \mathbf{U}_{\text{MSA}}^\top$.
    \item \textbf{Define Decoupling Loss:} During fine-tuning on the dialectal rewriting task, we add a penalty term to the standard language modeling loss. This \textit{decoupling loss} penalizes the magnitude of the projection of the model's hidden states $\mathbf{H}$ onto the MSA subspace:
    \begin{equation}
        \mathcal{L}_{\text{decouple}} = \mathbb{E}\left[\left\lVert \mathbf{H} \mathbf{P}_{\text{MSA}} \right\rVert_2 \right]
    \end{equation}
    The total loss is $\mathcal{L} = \mathcal{L}_{\text{LM}} + \lambda \, \mathcal{L}_{\text{decouple}}$, where $\lambda$ is a hyperparameter (we use 0.01).
\end{enumerate}
Crucially, the probe is periodically retrained on fresh model checkpoints during fine-tuning. This \textbf{online updating} of $\mathbf{P}_{\text{MSA}}$ ensures that our intervention targets the evolving MSA subspace, enabling a precise and adaptive causal manipulation of the model's representational geometry. Training details are in Appendix~\ref{sec:decouple-detailed}.

\subsection{Experimental Setup}
\paragraph{Data.} All experiments use the MADAR 25 corpus \citep{bouamor-etal-2018-madar}, which contains 2,000 parallel sentences across 25 city-level Arabic dialects, MSA, English, and French. This fine-grained, multi-dialect parallel resource is unique and enables our controlled study.

\paragraph{Models.} We analyze a suite of state-of-the-art open-weight multilingual models: Jais-family 30B \citep{sengupta2023jais}, Gemma 3 1B \citep{gemma_2025}, Aya expanse 8B \citep{dang2024ayaexpansecombiningresearch}, and Qwen 3 14B \citep{qwen3technicalreport}. For our causal intervention experiments, we deliberately select Gemma 3 1B. Its smaller parameter count implies a more constrained representational space, making it a challenging and informative test case for the benefits of explicit subspace management. Furthermore, its weaker baseline performance provides a clear opportunity to measure improvement from our method.

\section{Results and Analysis}
We now present our empirical investigation, which first diagnoses the representational pathologies hindering dialectal generation in multilingual models and then validates our hypothesis with a causal intervention.

\subsection{Baseline: Models Exhibit Poor Low-Resource Generative Capabilities}
\begin{wrapfigure}[21]{r}{0.5\textwidth}
    \includegraphics[width=\linewidth]{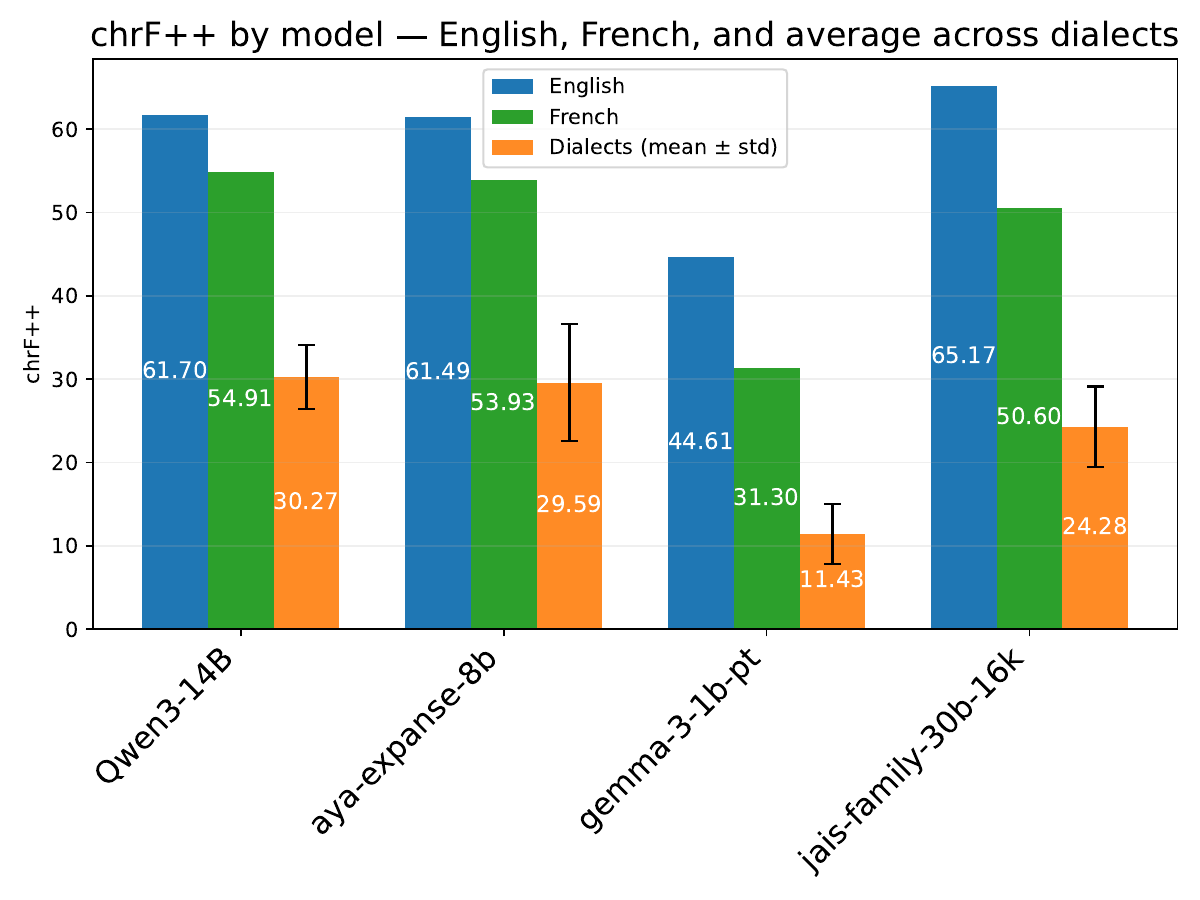}
        \caption{Generative performance on the dialectal rewriting task. All models perform poorly on low-resource dialects compared to high-resource languages, and performance does not correlate with model scale.}
        \label{fig:mt-results}
\end{wrapfigure}

We first benchmark the zero-shot performance of several models on our dialectal rewriting task. As shown in Figure~\ref{fig:mt-results}, all models struggle significantly, with even the best-performing models, Qwen 3 14B and Aya Expanse 8B, achieving only modest chrF++ scores. Notably, performance does not correlate with model scale or specialization; the largest, Arabic-centric Jais-family 30B model is outperformed by smaller models. This poor performance, especially when contrasted with the models' strong capabilities in high-resource language pairs (MSA to English/French), points to a more fundamental issue than a simple lack of capacity. We posit this stems from the model's internal representations.

\begin{figure}[]
    \includegraphics[width=\linewidth]{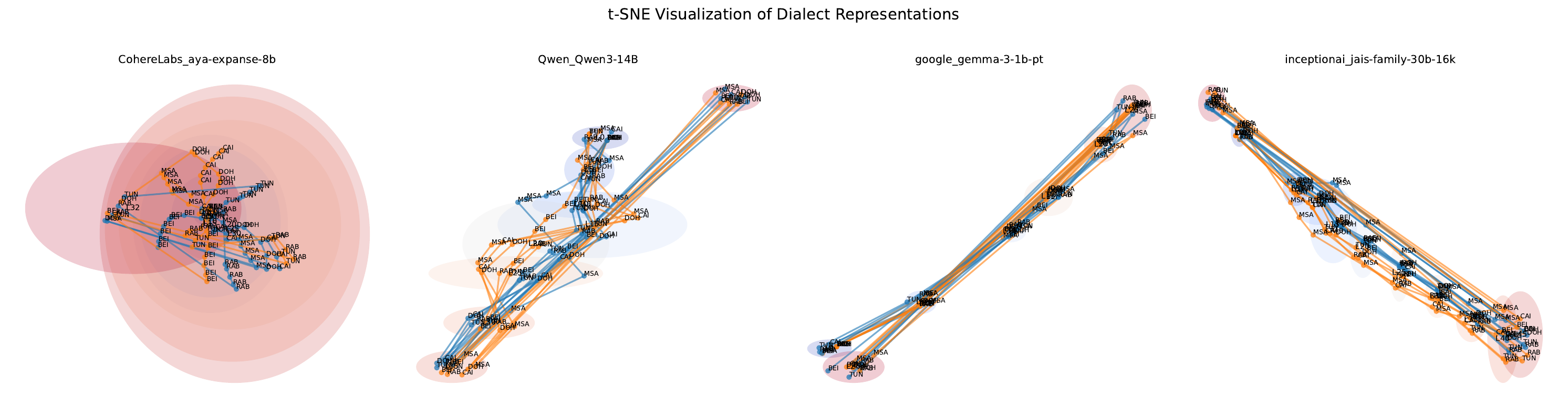}
        \caption{t-SNE of sentence representations. Higher-performing models (e.g., Qwen, Aya) exhibit clearer separation between dialectal clusters in their intermediate layers, unlike weaker models (Gemma, Jais).}
        \label{fig:tsne}
\end{figure}

\subsection{Diagnosis I: Geometric Analysis Links Performance to Representational Separation}
To investigate the underlying representational geometry, we visualize the hidden states of parallel sentences using t-SNE (Figure~\ref{fig:tsne}). The visualizations reveal a striking pattern: stronger models like Qwen and Aya learn to separate representations by dialect in their intermediate layers, whereas weaker models like Jais and Gemma maintain entangled representations. This qualitative observation suggests a link between a model's ability to geometrically isolate dialectal subspaces and its downstream generative performance.

\begin{figure*}[t]
    \centering
    \includegraphics[width=\linewidth]{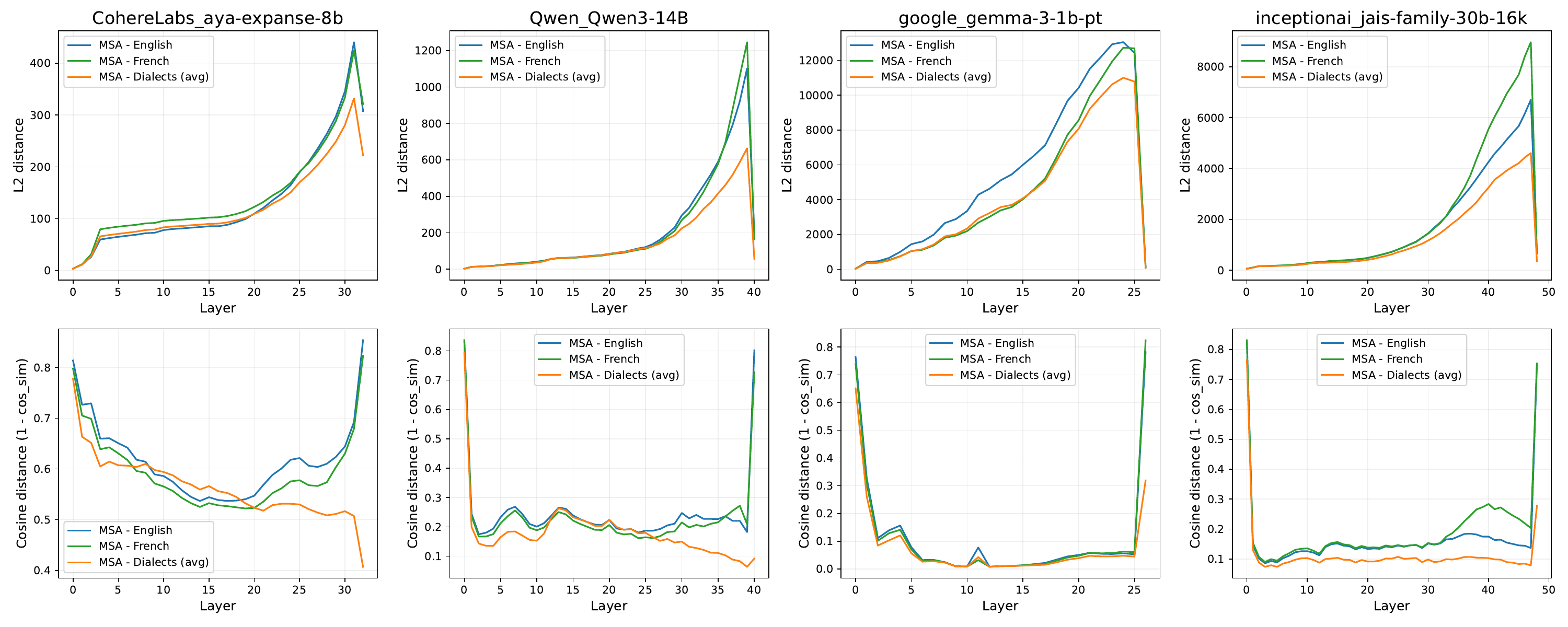}
    \caption{Layer-wise L2 (Top) and Cosine (Bottom) distance between dialectal representations and MSA. High-performing models show distinct geometric patterns, with Aya treating dialects more like separate languages (high L2 distance).}
    \label{fig:l2-distance-langs}
\end{figure*}

\begin{figure*}[t]
    \centering
    \includegraphics[width=\linewidth]{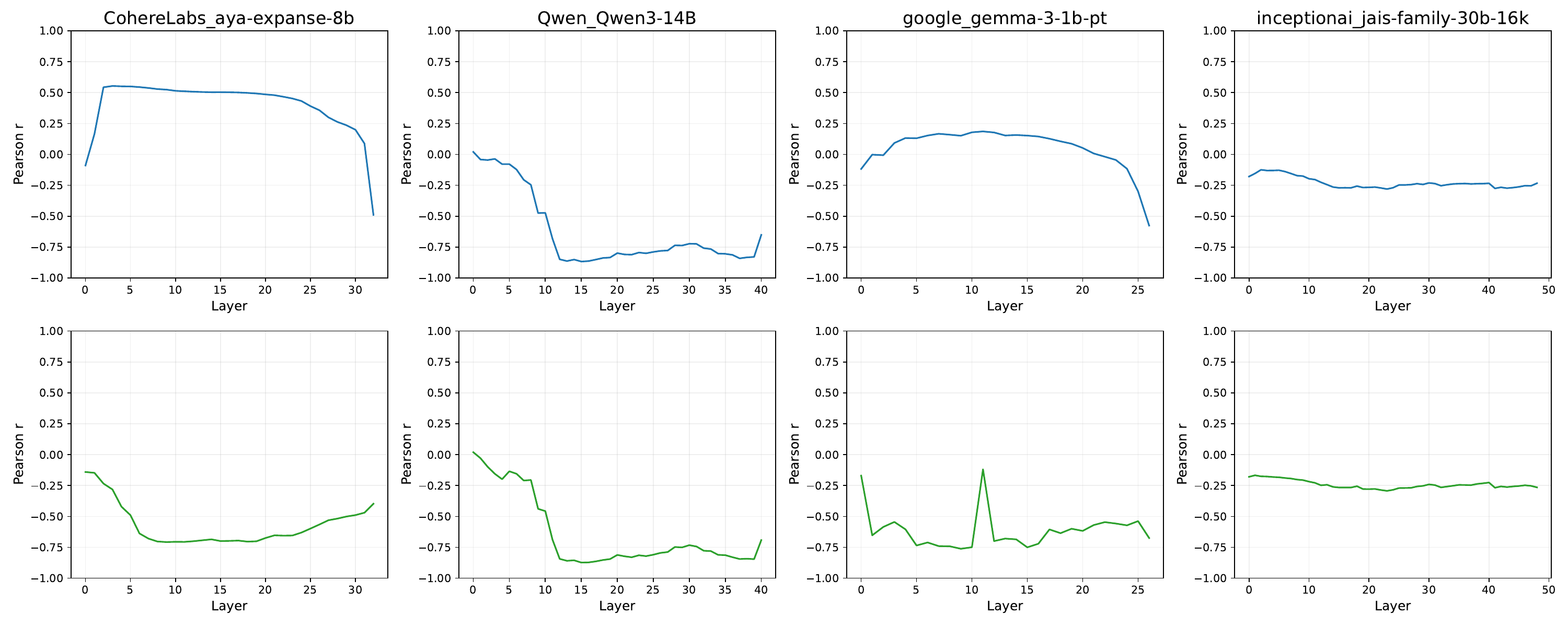}
    \caption{Layer-wise Pearson correlation between representational distance from MSA (L2-Top, Cosine-Bottom) and downstream generation performance. The consistent negative correlation with cosine distance suggests that subspace alignment is beneficial.}
    \label{fig:l2-dist-corr}
\end{figure*}

We quantify this by measuring the L2 and cosine distance between MSA and dialectal sentence representations across all layers (Figure~\ref{fig:l2-distance-langs}). We observe that different distance metrics capture different geometric properties: L2 distance reflects the degree of spatial separation, while cosine distance measures the alignment of subspaces. To substantiate the link to performance, we compute the layer-wise correlation between these distances and the chrF++ score (Figure~\ref{fig:l2-dist-corr}). A consistent negative correlation emerges between cosine distance and performance, especially in early-to-mid layers. This suggests that better alignment (lower cosine distance) in these layers is beneficial, likely facilitating the transfer of semantic information from the high-resource MSA. Conversely, the relationship with L2 distance is more complex, with models like Aya benefiting from greater spatial separation in intermediate layers. This indicates a delicate balance: subspaces must be aligned enough for knowledge transfer but separate enough to preserve unique dialectal features.

\subsection{Diagnosis II: Information-Theoretic Evidence of MSA's Representational Dominance}
\begin{wrapfigure}[15]{r}{0.5\textwidth}
    \includegraphics[width=\linewidth]{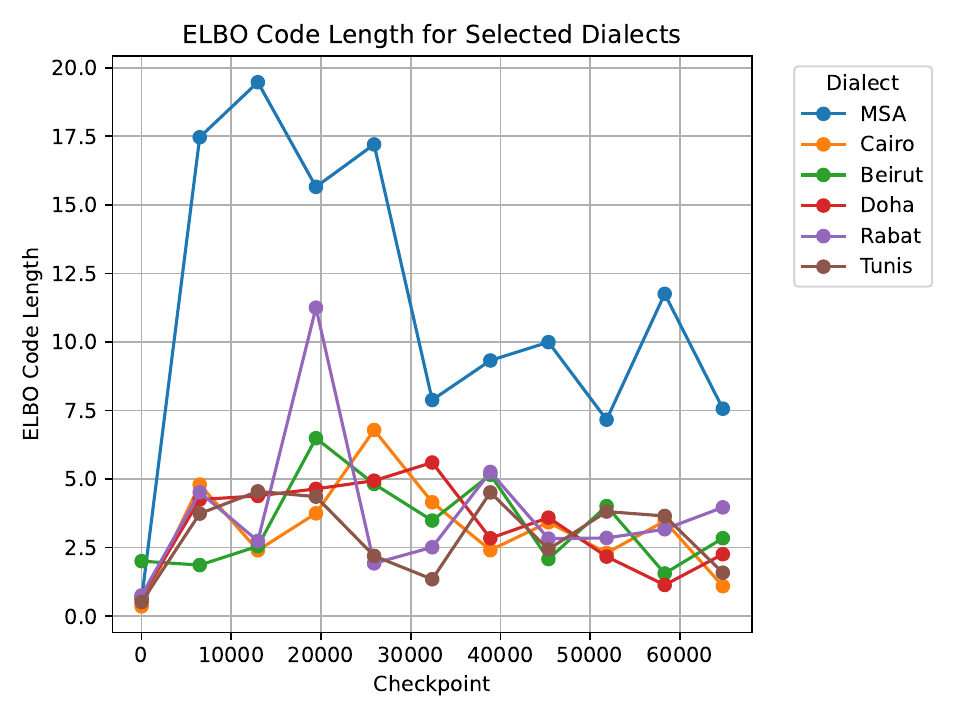}
        \caption{Code Length evolution over baseline training.}
        \label{fig:elbo_vs_time}
\end{wrapfigure}

The geometric analysis suggests entanglement with MSA is problematic. We further interrogate this using information-theoretic probing during standard supervised fine-tuning (SFT) on the dialectal rewriting task. We track the ELBO code length required to identify dialects from the model's hidden states (a proxy for how accessible this information is). As shown in Figure~\ref{fig:elbo_vs_time}, standard fine-tuning causes the code length for all dialects to increase slightly, as the model specializes for generation rather than classification. However, the increase is \textbf{disproportionately large for MSA}. This indicates that the model is actively making MSA-specific information less linearly accessible, suggesting its pre-trained MSA representation is oversized and detrimental to the dialectal generation task.

This ``pruning" of the MSA subspace has a direct geometric consequence. As we fine-tune, the Subspace Angle (SSA) between MSA and the dialectal subspaces consistently increases (Figure~\ref{fig:chrf_ssa}, left). That is, the dialectal subspaces systematically drift away from the MSA subspace. Crucially, this growing separation directly correlates with improvements in generation performance (Figure~\ref{fig:chrf_ssa}, right).

Taken together, these analyses provide compelling correlational evidence for our central hypothesis: the representational dominance of the high-resource standard language (MSA) actively hinders a model's ability to generate text in related low-resource varieties. Fine-tuning implicitly alleviates this by pushing dialectal representations away from the MSA subspace.

There are a few limitations to keep in mind of our analyses so far. Namely, that they have been observational and serve to build our hypothesis; the causal link is established specifically by the success of our decoupling intervention which we will discuss in Section~\ref{sec:online-decouple}. Finally, the MADAR dataset, while unique in its breadth of dialects, is composed of relatively short sentences. This setting may not fully capture model behaviors on tasks requiring longer-form generation, thereby defining the scope of our current findings. We hope future work addresses this gap in data availability.

\subsection{Causal Validation: Online Subspace Decoupling Boosts Performance}
\label{sec:online-decouple}
To move from correlation to causation, we test our hypothesis directly using our proposed \textbf{Online Subspace Decoupling} method (Section~\ref{sec:decouple}). By adding an explicit penalty term that pushes dialectal hidden states out of the MSA subspace, we actively enforce the representational separation that SFT appears to learn implicitly.

\begin{figure*}[t]
    \centering
    \includegraphics[width=\linewidth]{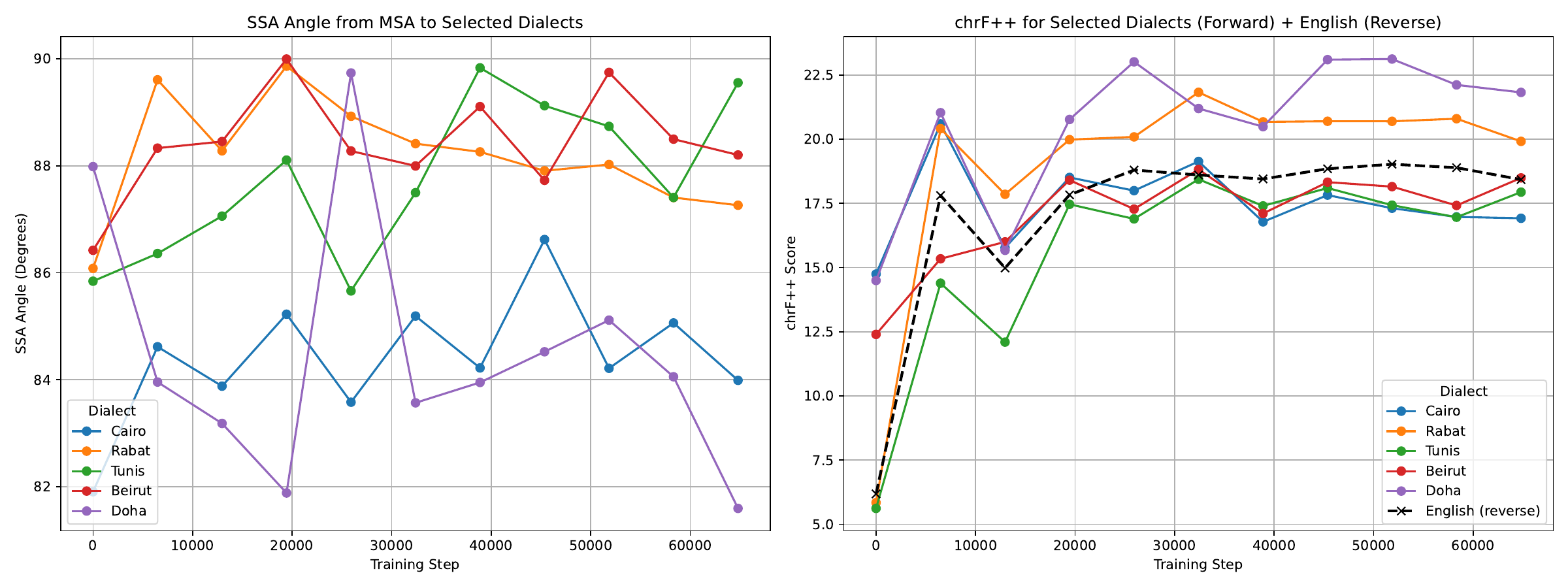}
    \caption{\textbf{(Left)} During SFT, the subspace angle (SSA) between MSA and dialects consistently increases, indicating growing representational separation. \textbf{(Right)} This increase in separation correlates directly with improved chrF++ scores. This provides strong evidence that disentangling from MSA is a key mechanism for improving dialectal generation.}
    \label{fig:chrf_ssa}
\end{figure*}

\begin{figure}
        \centering
        \includegraphics[width=0.5\linewidth]{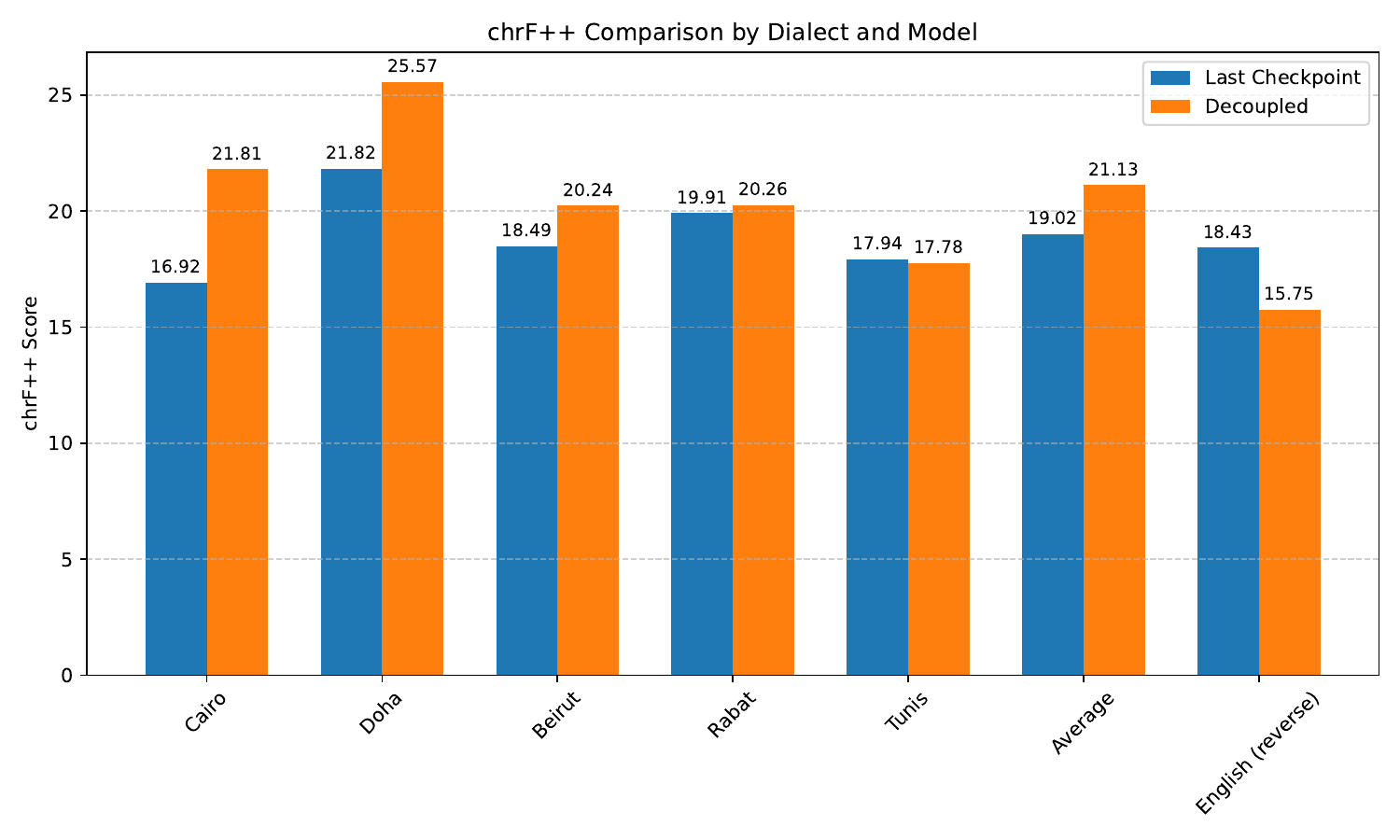}
        \caption{Our causal intervention (Online Decoupling, Orange) consistently improves performance over baseline SFT (Blue) by actively enforcing representational separation.}
        \label{fig:chrF_comparison}
\end{figure}
The results, shown in Figure~\ref{fig:chrF_comparison}, provide clear causal validation. Our intervention yields consistent and significant performance gains across nearly all dialects, achieving a \textbf{+2.0 chrF++ average improvement} and gains as high as \textbf{+4.89} for specific dialects (Cairo) over a standard SFT baseline. The dialects that benefit least (e.g., Tunis, Beirut) were already those with the highest initial separation from MSA (Figure~\ref{fig:chrf_ssa}), suggesting they were less affected by the entanglement problem. While there is an expected trade-off, performance on MSA generation drops, the substantial boost for a wide range of low-resource varieties confirms that mitigating representational dominance is a potent mechanism for improving generative capabilities. Visualizations (tSNE) in Appendix~\ref{sec:decouple-results-detailed} confirm that our method achieves a much greater degree of geometric separation than baseline SFT, directly linking the intervention to its intended structural effect on the model's internal representations.

Our analysis, while providing strong causal evidence, has several limitations that frame opportunities for future research. Our causal claim rests on intervention experiments within a single, albeit complex, language family: Arabic. While we hypothesize that the underlying mechanism of representational dominance is a general phenomenon, empirical validation on other language families is necessary to confirm this.

\section{Conclusion \& Future Work}
This work demonstrates that representational entanglement with a high-resource language is a critical and addressable bottleneck for generative modeling in closely-related, low-resource language varieties. Through a combination of geometric and information-theoretic analyses on Arabic dialects, we provided evidence that the representational dominance of Modern Standard Arabic (MSA) hinders dialectal generation. We then moved from correlation to causation, introducing a novel \textit{online subspace decoupling} method that actively and dynamically separates dialectal representations from the MSA subspace during fine-tuning.

Our experiments provide the first causal evidence that explicitly managing this subspace overlap yields substantial performance gains, up to +4.9 chrF++ on individual dialects and +2.0 on average, validating our hypothesis. While our method was designed for hypothesis testing and is computationally intensive, its success illuminates a clear path forward. The results highlight the critical importance of representational allocation in multilingual models and motivate future work in several key directions:

\begin{itemize}
    \item \textbf{Scalable and Efficient Methods:} Future research should focus on developing computationally cheaper alternatives that achieve similar decoupling effects. This includes designing parameter-efficient fine-tuning (PEFT) methods, such as subspace-aware adapters, or formulating novel pre-training objectives that encourage a more balanced representational space from the outset.
    
    \item \textbf{Inference-Time Interventions:} A particularly promising avenue is to move beyond training-based solutions. Inference-time techniques like activation steering or targeted model editing could offer a more surgical and efficient approach. For instance, identifying and patching neurons responsible for MSA-specific features could suppress the dominant language's influence on-the-fly, without requiring any gradient-based updates.
    
    \item \textbf{Generalization to Other Languages:} A crucial next step is to investigate whether these principles of representational entanglement generalize beyond Arabic. Applying our analytical framework and interventions to other language families with similar resource imbalances and orthographic overlap, such as the Czech-Slovak or Scandinavian language continuums, is essential for establishing the broader utility of subspace management in multilingual representation learning.
\end{itemize}

\subsubsection*{Acknowledgments}
This work was conducted during Ahmed Elshabrawy's research internship at NICT, Japan. We gratefully acknowledge the support and computational resources provided by NICT, Japan that made this research possible.

\bibliography{iclr2026_conference}
\bibliographystyle{iclr2026_conference}

\appendix
\section{Tokenizer Fertilities}
\label{sec:tokenizers}
To check how dialects are handled differently to MSA at the most basic level by models we examine various tokenizer fertilities. We include tokenizers of all the model families we explore. Furthermore, we train custom WordPiece tokenizers with an 8,000 vocabulary size \citep{wu-wordpiece} on the train and dev set of the MADAR 25 corpus of 6 different dialects/varieties (each having 10,000 sentences in total). 

\begin{figure*}
    \centering
    \includegraphics[width=\linewidth]{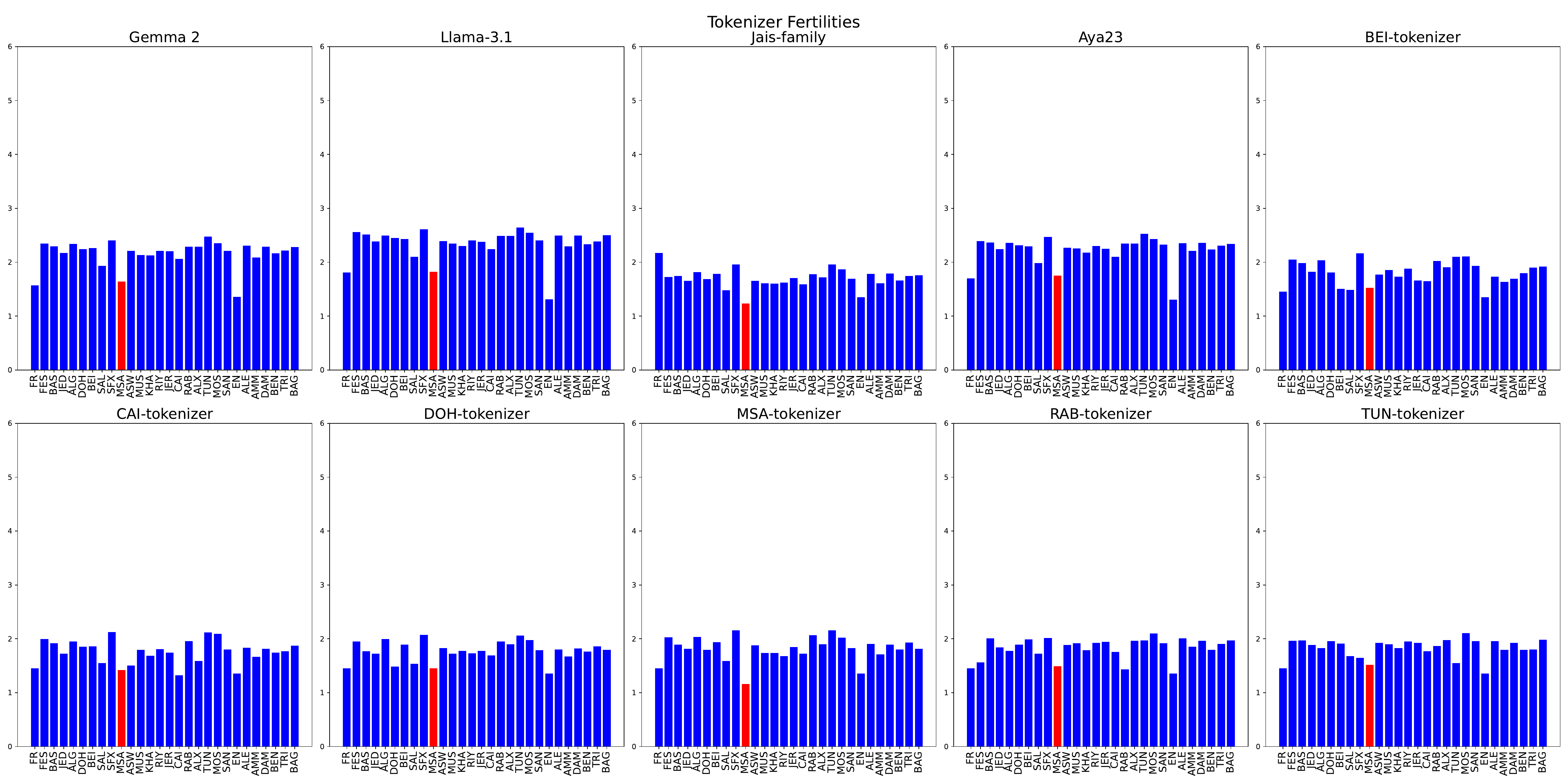}
    \caption{Tokenizer fertilities of various language model families, as well as a WordPiece tokenizer trained from scratch on a specifc dialectal set. The red bar corresponds to MSA.}
    \label{fig:tokenizers}
\end{figure*}

As seen in Figure~\ref{fig:tokenizers}, there are several interesting observations. First, it is clear that there is a divide between MSA, English, and French and the non-standard dialects of Arabic which have significantly higher fertilities. 

Furthermore, Across \textbf{all} tokenizers, dialects tend to have much higher fertilities to MSA. Interestingly, even for tokenizers trained on a single dialect, the fertility of MSA tends to be close or even lower than the dialect it was trained on. This seems to indicate that MSA serves a lowest common denominator for Arabic tokenizers whereas dialects can have more foreign/disjoint vocabularies (i.e. most dialectal varieties will feature a minimum MSA lexicon with other elements which are not shared). This is consistent with the linguistic intuition and is supported by the unique word counts of the train sets of all the dialectal tokenizers show below:
\begin{table}
    \centering
    \begin{tabular}{lc}
        \toprule
        Dialect & \#Unique Words \\
        \midrule
        DOH  & 12,651 \\
        BEI  & 15,083 \\
        MSA  & 12,197 \\
        CAI  & 14,611 \\
        RAB  & 16,136 \\
        TUN  & 15,437 \\
        \bottomrule
    \end{tabular}
    \caption{Unique words}
    \label{tab:data}
\end{table}

Interestingly, we note however that training on a single dialect seems to lead to lower fertilites on all other dialects. Especially when comparing to the LLM and MSA-only tokenizers observed. This is indicative that a majority of the training data used to train the multilingual tokenizers of all models observed is lacking in dialectal data.

This highlights that before any processing is done by models dialectal words are inherently treated differntly at the lowest stage of tokenization. Furthermore, training tokenizers on MSA data only is not sufficient for tokenizers to capture dialectal words sufficiently. 

\section{City Names to Dialect Code}
\begin{table}
    \centering
    
    \begin{tabular}{l|l}
        \textbf{City} & \textbf{Code} \\
        \hline
        Rabat & RAB \\
        Fes & FES \\
        Algiers & ALG \\
        Tunis & TUN \\
        Sfax & SFX \\
        Tripoli & TRI \\
        Benghazi & BEN \\
        Cairo & CAI \\
        Alexandria & ALX \\
        Aswan & ASW \\
        Khartoum & KHA \\
        Jerusalem & JER \\
        Amman & AMM \\
        Salt & SAL \\
        Beirut & BEI \\
        Damascus & DAM \\
        Aleppo & ALE \\
        Mosul & MOS \\
        Baghdad & BAG \\
        Basra & BAS \\
        Doha & DOH \\
        Muscat & MUS \\
        Riyadh & RIY \\
        Jeddah & JED \\
        Sana’a & SAN \\
        \hline
    \end{tabular}
    \caption{City Names and Their Codes}
    \label{tab:city_codes}
\end{table}

\section{More Information About Probing}
\label{sec:probing-detailed}
To complement geometric subspace analysis, we adopt an information-theoretic variational linear probe \citep{voita-titov-2020-information, muller-eberstein-etal-2023-subspace} to quantify how much dialect identity information is recoverable from token-level model representations. For a given token, let $\{\mathbf{h}^{(0)}, \dots, \mathbf{h}^{(\ell)}\} \in \mathbb{R}^d$ denote its hidden states from all $\ell$ layers, including the non-contextualized layer $0$. The probe computes a learned weighted average over layers:
\[
\mathbf{h}' = \sum_{i=0}^\ell \alpha_i \mathbf{h}^{(i)},
\]
where $\boldsymbol{\alpha} \in \mathbb{R}^\ell$ are learned combination weights.

This aggregated representation is fed to a linear classifier with weight matrix $\boldsymbol{\theta} \in \mathbb{R}^{d \times c}$ for $c$ dialect classes. Following \citet{voita-titov-2020-information}, each weight $w$ in $\boldsymbol{\theta}$ is drawn from a normal distribution
\[
w \sim \mathcal{N}(z\mu, z^2\sigma^2),
\]
where the scaling factor $z$ is also drawn from
\[
z \sim \mathcal{N}(\mu_z, \sigma^2_z).
\]
The pair $(w,z)$ is given a joint normal--Jeffreys prior
\[
\gamma(w,z) \propto |z|^{-1} \, \mathcal{N}(w \mid 0, z^2)
\]
which encourages sparsity by pushing weights toward zero with low variance.

The probe parameters $(\boldsymbol{\alpha}, \boldsymbol{\theta})$ are trained to minimize
\[
\mathcal{L} = \mathrm{CE}(y, \hat{y}) \;+\; \beta \, D_{\mathrm{KL}}\!\left( q(\boldsymbol{\theta}) \,\|\, \gamma(\boldsymbol{\theta}) \right),
\]
where $\mathrm{CE}$ is the cross-entropy loss for one-vs-rest dialect classification, and the KL term regularizes $\boldsymbol{\theta}$ toward the sparsity-inducing prior. This objective maximizes compression while preserving predictive accuracy, yielding a layer-combined, token-level estimate of recoverable dialect identity information. The one-vs-rest objective hones in on dialect specific information that can help the model discern between similar dialects and offers counter-examples. We construct the training set for each dialect/variety/language by taking all the target's sentences in MADAR 26's training set, we construct an equal number of counter-examples from all the other dialects and languages. We make this data available (anonymized). We include training hyperparameters for the probes in Table~\ref{tab:probe-hyperparams}.

\begin{table}[h]
\centering
\resizebox{0.5\linewidth}{!}{
\begin{tabular}{ll}
\hline
\textbf{Hyperparameter} & \textbf{Value} \\
\hline
Model name & \texttt{google/gemma-3-1b-pt} \\
KL weight & 1.0 \\
Number of epochs & 30 (for analysis) \\
 & 15 (for decoupling training) \\
Early stopping patience & 5 \\
\hline
\end{tabular}
}
\caption{Training hyperparameters for variational probe experiments.}
\label{tab:probe-hyperparams}
\end{table}

\section{Online Decoupling Training Details}
\label{sec:decouple-detailed}
This appendix outlines the key design decisions underlying our online MSA subspace decoupling method, as well as the exact hyperparameters used in our experiments. 

\subsection{Design Choices}

\paragraph{Projection Subspace Estimation.}
We estimate the MSA subspace using a \textit{variational linear probe} trained on a Modern Standard Arabic (MSA) vs.\ non-MSA dialect identification task over the MADAR corpus. We recover the subspace basis from the learned probe parameters using Singular Value Decomposition (SVD) of the parameter matrix $\theta_{\text{MSA}}$. The number of retained singular vectors equals the probe’s latent dimension.

\paragraph{Online Updating.}
Rather than estimating the MSA subspace once before training, we periodically retrain the probe on the current model checkpoint during fine-tuning. This ensures that the projection matrix $\mathbf{P}_{\text{MSA}}$ remains synchronized with the evolving hidden representation geometry. The projection matrix is updated every $N_{\text{update}}$ gradient steps.

\paragraph{Layer Aggregation.}
Hidden representations from all layers are combined using a learned set of attention weights $\alpha \in \mathbb{R}^{L+1}$ from the variational probe. This allows the method to focus the decoupling penalty on layers most predictive of MSA features.

\paragraph{Penalty Formulation.}
We penalize the $\ell_2$ norm of the projection of the aggregated hidden states onto the MSA subspace:
\begin{equation}
\mathcal{L}_{\text{decouple}} = \mathbb{E} \left[ \left\lVert \mathbf{H} \mathbf{P}_{\text{MSA}} \right\rVert_2 \right],
\end{equation}
where $\mathbf{H}$ are the contextual hidden states and $\mathbf{P}_{\text{MSA}}$ is the projection matrix.

\paragraph{Loss Weighting.}
The decoupling penalty is scaled by a coefficient $\lambda$ and added to the standard causal language modeling loss:
\begin{equation}
\mathcal{L} = \mathcal{L}_{\text{LM}} + \lambda \cdot \mathcal{L}_{\text{decouple}}.
\end{equation}

\paragraph{Bidirectional Training Data.}
To encourage symmetric modeling of both MSA~$\rightarrow$~dialect and dialect~$\rightarrow$~MSA directions, we construct bidirectional rewriting prompts for each sentence pair.

\subsection{Hyperparameters}

\begin{table*}[h]
\centering
\begin{tabular}{ll}
\toprule
\textbf{Parameter} & \textbf{Value / Setting} \\
\bottomrule
Base model & \texttt{google/gemma-3-1b-pt} \\
Tokenizer & Matching HF tokenizer (\texttt{pad\_token = eos\_token}) \\
Batch size (per device) & 1 \\
Gradient accumulation steps & 4 \\
Max sequence length & 512 \\
Optimizer & AdamW (via HF Trainer default) \\
Learning rate & $5\times10^{-5}$ (default HF schedule) \\
Loss coefficient $\lambda$ & 0.01 \\
Probe update steps $N_{\text{update}}$ & 500 \\
Probe training epochs & 15 \\
Probe dataset & anonymous dataset (DID-MSA) \\
Probe input type & Sequence-level dialect identification \\
Number of probe classes & 2 (MSA vs.\ non-MSA) \\
Projection estimation & SVD on $\theta_{\text{MSA}}$ \\
Subspace dimensionality & Full rank of $\theta_{\text{MSA}}$ \\
Layer aggregation & Learned attention weights $\alpha$ \\
Early stopping patience & 3 epochs (validation loss) \\
Early stopping threshold & 0.01 \\
Train/validation split & 90\% / 10\% \\
\bottomrule
\end{tabular}
\caption{Hyperparameters used in online MSA decoupling experiments.}
\label{tab:hyperparams-decoupling}
\end{table*}

\section{Decoupling Results for All Dialects}
\label{sec:decouple-results-detailed}

\begin{figure*}
    \centering
    \includegraphics[width=\linewidth]{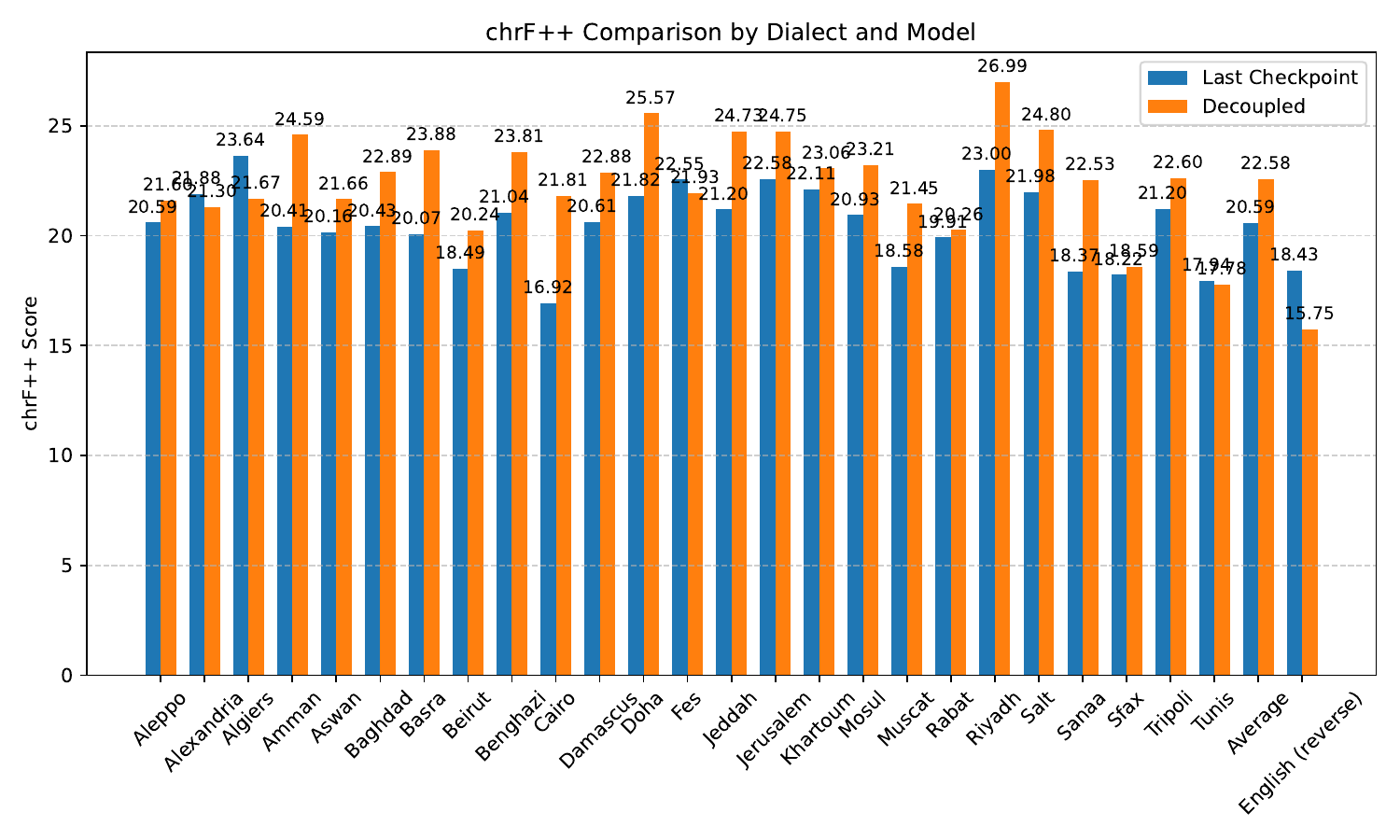}
    \caption{Baseline ChrF++ (Blue) vs. our novel online decoupling method (Orange) on Multiple dialects and on English to MSA translation.}
    \label{fig:chrF_comparison_25}
\end{figure*}

In Figure~\ref{fig:chrF_comparison_25}, we show the comparison of the baseline SFT and the decoupling method for all 25 dialects. The overwhelming majority of dialects benefit from our novel training. Aleppo, Alexandria, Fes, Rabat, Sfax, and Tunis seem to benefit the least from the method seeing barely any changes when compared to baseline SFT. Algiers seems to be the only dialect (aside from MSA) that is actively hurt from the intervention. It is difficult to explain why exactly it stands out; however, given that the overwhelming majority dialects seem to benefit from the intervention it seems to be overall beneficial to DiaMT.

\section{tSNE post SFT Training}

\begin{figure*}
    \centering
    \includegraphics[width=\linewidth]{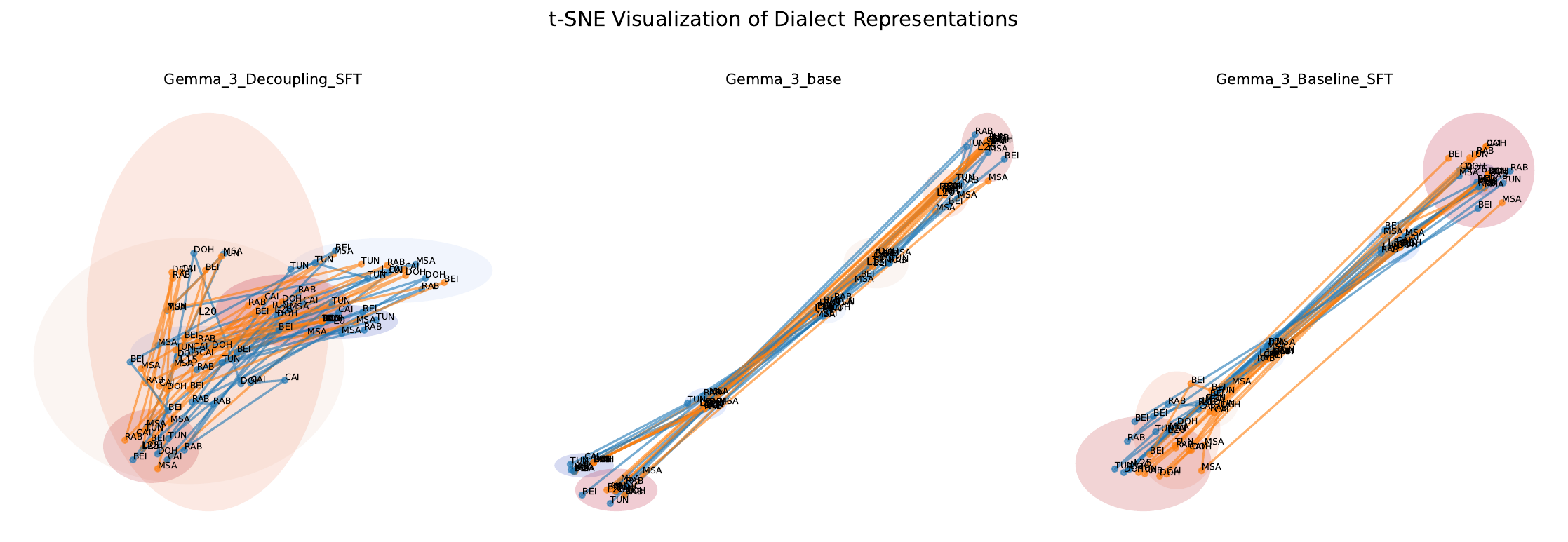}
    \caption{Visualizing internal model representations every 5 layers across 2 different sentences (Sentence 0 in blue and Sentence 1 in orange) written in 6 varieties of Arabic using t-SNE. Layer spaces are approximated with ellipses with a color gradient from layer 0 (blue) to the last layer (red). Left is our novel decoupling training, center is the base model, right is the baseline SFT training on DiaMT.}
    \label{fig:tsne-comp}
\end{figure*}

\end{document}